\documentclass[final]{l4dc2026}

\title[Embodied Learning of Reward]{Embodied Learning of Reward for Musculoskeletal Control \\ with Vision Language Models}

\usepackage{times}
\usepackage{amsfonts}
\usepackage{amsmath}
\usepackage{amssymb}
\usepackage{mathtools}
\usepackage{algorithm}
\usepackage{algcompatible}
\usepackage{eqparbox}
\usepackage{wrapfig}

\usepackage{hyperref}
\usepackage{url}
\usepackage[colorinlistoftodos]{todonotes}
\usepackage{verbatim}
\usepackage{subcaption}
\usepackage{caption}
\newcommand{\algo}{\textit{MoVLR}} 
\newcommand{\mpc}{MPC$^2$}

\usepackage{listings}
\usepackage{xcolor}
\usepackage{pdfpages}
\usepackage{wrapfig}

\author{%
 \Name{Saraswati Soedarmadji}$^1$ \Email{chenxuying24@mails.tsinghua.edu.cn}\\
 \Name{Yunyue Wei}$^1$ \Email{yunyuewei@mail.tsinghua.edu.cn}\\
 \Name{Chen Zhang}$^1$ \Email{czhang.email@gmail.com}\\
 \Name{Yisong Yue}$^2$ \Email{yyue@caltech.edu}\\
 \Name{Yanan Sui}$^1$ \Email{ysui@tsinghua.edu.cn}\\
 \addr $^1$Tsinghua University, $^2$California Institute of Technology%
}

\begin{document}

\maketitle

\begin{abstract}
Discovering effective reward functions remains a fundamental challenge in motor control of high-dimensional musculoskeletal systems.
While humans can describe movement goals explicitly such as ``walking forward with an upright posture,'' the underlying control strategies that realize these goals are largely implicit, making it difficult to directly design rewards from high-level goals and natural language descriptions. 
We introduce Motion from Vision-Language Representation (\algo), a framework that leverages vision-language models (VLMs) to bridge the gap between goal specification and movement control.
Rather than relying on handcrafted rewards, \algo\ iteratively explores the reward space through iterative interaction between control optimization and VLM feedback, aligning control policies with physically coordinated behaviors. Our approach transforms language and vision-based assessments into structured guidance for embodied learning, enabling the discovery and refinement of reward functions for high-dimensional musculoskeletal locomotion and manipulation. These results suggest that VLMs can effectively ground abstract motion descriptions in the implicit principles governing physiological motor control.

\end{abstract}


\section{Introduction}
\label{sec:introduction}

Humans acquire motor control through practice, imitation, and external guidance, with effective control arising from the intricate interactions between the nervous and musculoskeletal systems. Unlike general robotic systems, musculoskeletal agents exhibit highly nonlinear, overactuated, and high-dimensional dynamics, where multiple muscles or synergies can produce identical joint motions. Coordinated movement therefore depends on discovering appropriate patterns of whole-body control rather than specifying individual actuator commands, making the realization of efficient and natural motion fundamentally challenging. 

Learning-based approaches have enabled progress in high-dimensional musculoskeletal control~\citep{MyoSuite2022, schumacher2023dep, he2024dynsyn}. However, most existing methods rely on heuristic objectives such as velocity tracking or energy minimization, which often fail to capture the nuanced structure of motion complexity. While sufficient for basic task completion, such objectives often neglect anatomical principles and lead to biomechanically unnatural or inefficient behaviors.

Recent advances in large language models (LLMs) and vision-language models (VLMs) suggest a promising direction for exploring reward structure through high-level assessments of motion quality and coordination~\citep{sontakke2023roboclip, ma2023eureka,zeng2024video2reward}. Nevertheless, existing approaches typically rely on episodic statistics or coarse success signals as feedback and are evaluated primarily on low-dimensional, torque-driven systems with explicit dynamics. Motor control, however, is governed by temporally extended and implicitly structured sensorimotor dynamics that are difficult to capture through such feedback alone. As a result, it remains unclear whether these models can systematically translate implicit interaction dynamics into structured rewards for high-dimensional musculoskeletal control.

\begin{figure}[htp]
    \centering
    \includegraphics[width=1\linewidth]{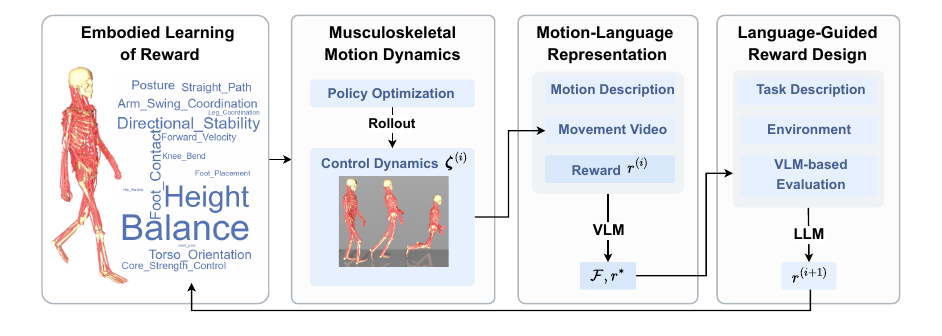}
    \caption{Workflow of \algo.
    Policy optimization is performed to provide high-dimensional musculoskeletal dynamics of the reward candidate. A VLM evaluates the corresponding movement video $\boldsymbol{\zeta}^{(i)}$ to update the current best reward design $r^*$ and suggest biomechanical improvements $\mathcal{F}$ for a LLM to refine reward generation of $r^{(i+1)}$.
    }
    \label{fig:pipeline}
\end{figure}

In this paper, we introduce Motion from Vision Language Representation (\algo), a framework for automatic reward learning in high-dimensional musculoskeletal systems that integrates descriptive and dynamical feedback. As shown in Figure~\ref{fig:pipeline}, \algo~extracts musculoskeletal dynamics through policy optimization and rollout, rendering the resulting behaviors as movement videos. A vision-language model evaluates these motions to produce structured biomechanical feedback, which is then used by a language model to refine the reward formulation in subsequent iterations. Through this iterative process, implicit dynamical regularities are progressively distilled into explicit reward terms. By grounding temporal dynamics in semantically meaningful descriptors, \algo~bridges perceptual feedback with domain-informed motion representations, enabling scalable and biomechanically realistic reward discovery for high-dimensional musculoskeletal control.
The supplementary material and experiment demonstrations can be found at: \href{https://lnsgroup.cc/research/MoVLR/}{https://lnsgroup.cc/research/MoVLR/}.

\textbf{Contributions:} (1) We present \algo, a fully automatic framework for discovering reward structure in high-dimensional musculoskeletal systems by capturing implicit dynamics and refining them into explicit, executable rewards for control. (2) We demonstrate that \algo\ generalizes across movement tasks, environments, and morphologies, producing interpretable reward terms that reflect underlying musculoskeletal dynamics. (3) We introduce a unified framework that enables interpretable evaluation of motor performance, adaptive reward refinement through vision–language feedback, and transferable control for natural and coordinated motion.

\section{Related Works}
\label{sec:related}

\subsection{Control of musculoskeletal systems}

The control of high-dimensional musculoskeletal systems remains challenging due to the redundancy and dimensionality of human-like actuation. Significant progress has been made in developing faithful simulation environments that model muscle–tendon dynamics and joint kinematics, enabling more realistic learning and evaluation \citep{lee2019scalable,song2021deep,MyoSuite2022}. To address control complexity, prior work has explored hierarchical decompositions \citep{lee2019scalable,park2022generative,feng2023musclevae}, curriculum learning strategies \citep{caggiano2023myodex,park2025magnet}, bio-inspired sampling \citep{schumacher2023dep}, latent-space coordination \citep{chiappa2024latent}, and model-based control \citep{hansen2023td}. More recent approaches further reduce effective dimensionality by extracting muscle synergies informed by anatomy or task structure \citep{berg2024sar,he2024dynsyn}.

Despite these advances, achieving natural and human-like behaviors remains difficult. Performance is highly sensitive to reward design, where small changes can lead to unnatural postures, inefficient coordination, or brittle behaviors. Crafting such rewards typically requires domain expertise and manual tuning, and often relies on indirect proxies rather than direct biomechanical measures of movement quality~\citep{song2021deep, sui2017quantifying}. This reliance on hand-crafted objectives motivates the development of more expressive and adaptive reward design mechanisms, naturally leading to language and multimodal-driven approaches. 

\subsection{Language and multimodal driven reward design}  

Recent advances in large language models have shown strong potential for facilitating reward and feedback design in robotics and simulation systems \citep{goyal2019using, ma2024dreureka, yu2023language}. Eureka~\citep{ma2023eureka}, for example, uses code-generating LLMs to synthesize dense reward functions that exceed manually engineered counterparts in expressivity and task relevance. Complementary to this, language-conditioned reward modulation has also been explored as a mechanism for shaping policy learning during pretraining, enabling more flexible and semantically aligned optimization objectives \citep{adeniji2024languagereward}. Although originally introduced for reinforcement learning, this paradigm is equally applicable to robotics control, where generated signals can be interpreted as structured feedback shaping system dynamics toward desired trajectories~\citep{brohan2023rt1, driess2023palmE}.

Beyond text-only models, recent work with vision–language models demonstrates the benefits of incorporating multimodal inputs such as images or video into feedback design \citep{rocamonde2023vision, zitkovich23a, ge2023policy, wang2024rl, zeng2024video2reward}. For example, HARMON~\citep{jiang2024harmon} leverages a VLM to iteratively refine humanoid motion by evaluating rendered frame sequences against language descriptions. More recent work uses VLMs to detect and reason about failure modes in robotic manipulation, enabling richer evaluative feedback to inform reward design \citep{duan2025aha}. Despite these advances, existing approaches lack a principled framework for transforming LLM and VLM feedback into structured dynamical signals that directly shape reward functions, instead relying on heuristic or loosely coupled representations.


\section{Preliminaries}

\subsection{High-dimensional musculoskeletal control}
\textbf{Musculoskeletal systems.} In this paper, our target systems are high-dimensional, over-actuated musculoskeletal systems with dynamics governed by  
\begin{equation}
\boldsymbol{M(\boldsymbol{q})}\,\ddot{\boldsymbol{q}} + \boldsymbol{c}(\boldsymbol{q},\dot{\boldsymbol{q}}) = \boldsymbol{J}_m^\top \boldsymbol{f}_m + \boldsymbol{J}_c^\top \boldsymbol{f}_c + \boldsymbol{\tau}_{\mathrm{ext}},
\end{equation}
where $\boldsymbol{q}$ are generalized joint coordinates, $\boldsymbol{M(\boldsymbol{q})}$ is the mass matrix, and $\boldsymbol{c}(\boldsymbol{q},\dot{\boldsymbol{q}})$ represents Coriolis and gravitational effects. The Jacobians $\boldsymbol{J}_m$ and $\boldsymbol{J}_c$ map actuator and constraint forces ($\boldsymbol{f}_m, \boldsymbol{f}_c$) to generalized coordinates, and $\tau_{\mathrm{ext}}$ denotes external torques from the environment. Muscles are modeled as first-order actuators driven by neural controls $\boldsymbol{u}$ with activation $\boldsymbol{a}$, where the force $f_m$ generated by one actuator is formulated by:
\begin{equation}
f_m = F_k(l,v)\,a + F_p(l), 
\quad 
\frac{\partial a}{\partial t} = \frac{u - a}{\tau(u, a)},
\end{equation}
with actuator length $l$, velocity $v$,gains $F_k$, bias $F_p$ and time coefficient $\tau$. Note that $F_k, F_p$ and $\tau$ vary with muscle states, leading to high non-linearity. In our experiments, we use MS-Human-700 model~\citep{zuo2024self} as the major benchmark for human full-body musculoskeletal control. The model consists of $206$ joints and $700$ muscle-tendon actuators. Additional experiments can involve other morphologies.

\noindent\textbf{Policy optimization problem.} We model the high-dimensional musculoskeletal control problem as a finite horizon Markov decision process (MDP) with state $\boldsymbol{s}\in\mathcal{S}$, control $\boldsymbol{u}\in\mathcal{U}$, dynamics $\boldsymbol{f}\coloneqq\mathcal{S}\times\mathcal{U}\rightarrow\mathcal{S}$, horizon $T$, policy $\boldsymbol{\pi}\coloneqq \mathcal{S}\rightarrow\mathcal{U}$ and reward $r\coloneqq \mathcal{S}\times\mathcal{U}\rightarrow\mathbb{R}$. In
this paper, we formalize the reward as linear combination of reward terms:
$r = \boldsymbol{w}\cdot \boldsymbol{r}$ with $\boldsymbol{w}=(w_1, w_2, \cdots)$ and $\boldsymbol{r}=(r_1, r_2, \cdots)$ as the weights and values of specific reward terms. For example, a reward for human walking can be expressed as:
\begin{align}
    r_{\text{walk}} = w_{\text{height}}\cdot r_{\text{height}} + w_{\text{balance}}\cdot r_{\text{balance}} + \cdots + w_{\text{forward}}\cdot r_{\text{forward}}.
\end{align}
Given initial state $\boldsymbol{s}_0$, we aim to achieve stable control of the system by finding a policy $\boldsymbol{\pi}^*$ that maximize the reward function:
\begin{align}
\label{eq:policy_opt}
    \boldsymbol{\pi}^* = \text{argmax}_{\boldsymbol{\pi}}\sum_{t=0}^{T-1} r(\boldsymbol{s}_t, \boldsymbol{u}_t), \quad \boldsymbol{u}_t=\boldsymbol{\pi}(\boldsymbol{s}), \quad \boldsymbol{s}_{t+1} = \boldsymbol{f}(\boldsymbol{s}_t, \boldsymbol{u}_t).
\end{align}
The above policy optimization problem is also equivalent to reward minimization commonly used in control-based methods, where the reward function is the negative reward.

\subsection{Reward learning for musculoskeletal control}

While the above control problem provides single-step reward definition, the control performances are usually evaluated over full horizon. The objective of reward learning is to find single step reward $r^*$ that maximize the global reward function $R$:
\begin{align}
    r^* = \text{argmax}_r R(\boldsymbol{\zeta}), \quad \boldsymbol{\zeta}=(\boldsymbol{s}_0, \boldsymbol{a}_0,\cdots,\boldsymbol{s}_{T-1}, \boldsymbol{a}_{T-1}),\quad \boldsymbol{u}_t=\boldsymbol{\pi}^*_r(\boldsymbol{s}_t), \quad \boldsymbol{s}_{t+1} = \boldsymbol{f}(\boldsymbol{s_t}, \boldsymbol{u}_t),
\end{align}
where $\boldsymbol{\pi}^*_r$ is the optimal policy derived by maximizing $r$. 
In practice, the global reward $R$ is specified at a high level using natural language descriptions such as ``walk forward'' or ``grasp the bottle.'' The single-step reward $r$ consists of multiple reward terms and parameters as code pieces which need to be compatible with the policy optimization framework. Effective reward learning requires two key components: (1) \textbf{Embodied understanding of human movement} which extracts implicit biomechanical knowledge from motion descriptions, and (2) \textbf{Effective reward design} which integrates multimodal feedback to produce interpretable and executable reward terms for policy optimization.

  \begin{wrapfigure}{R}{0.55\textwidth}
    \begin{minipage}{0.55\textwidth}
\begin{algorithm}[H]
\caption{\algo}
\label{alg:pseudo_algo}
\begin{algorithmic}[1]
\REQUIRE Motion description $R$, environment code $\mathcal{E}$, max iterations $N$, initial reward design $r^{(0)}$
\smallskip
\STATE  $\boldsymbol{\zeta}^{*} \gets \emptyset$, $r^*\gets \emptyset$
\smallskip
\FOR{$i = 0, \cdots, N-1$}
    \STATEx
\COMMENT{\texttt{Musculoskeletal motion dynamics}}
    \STATE Obtain $\boldsymbol{\pi}^*_{r^{(i)}}$ by optimizing e.q. (\ref{eq:policy_opt})
    \STATE Obtain $\boldsymbol{\zeta}^{(i)}$ by rollout $\boldsymbol{\pi}^*_{r^{(i)}}$ 
    \STATEx\COMMENT{\texttt{Motion-language representation}}
    \STATE $\boldsymbol{\zeta}^*, r^* \gets \text{VLM}(R, \boldsymbol{\zeta}^{(i)},\boldsymbol{\zeta}^*, r^{(i)}, r^*)$
    \STATE $ \mathcal{F}\gets \text{VLM}(R,\boldsymbol{\zeta}^*, r^*)$
    \STATEx\COMMENT{\texttt{Language-guided reward design}}
    \STATE
        $r^{(i+1)} \sim \text{LLM}(R, \mathcal{E}, \mathcal{F}, r^{*})$ 
\ENDFOR
\STATE \textbf{Return:} Optimized reward $r^*$
\end{algorithmic}
\end{algorithm}
\end{minipage}
\end{wrapfigure}
 
\section{Methods}
To address the challenges of limited task understanding and multimodal reasoning, we propose \algo, a control-in-the-loop framework that integrates large language models (LLMs) and vision–language models (VLMs) into the reward learning process. \algo~bridges \textbf{explicit behavior specifications} expressed in natural language with the \textbf{implicit dynamical representations} essential for effective control. The key idea is to incorporate video observations of policy-executed trajectories into the iterated learning loop, enabling the model to jointly reason over linguistic intent and physical motion. This multimodal feedback provides structured insights into trajectory feasibility, biomechanical coherence, and task completion, ultimately yielding reward functions that are more aligned with the underlying system dynamics.

The workflow of \algo~is summarized in Algorithm \ref{alg:pseudo_algo}. At each iteration, the policy is optimized based on the current reward proposal, producing dynamical feedback that reflects the control performance (\textbf{musculoskeletal motion dynamics}, line 3-4). Given the executed control dynamics (rendered as video), a vision–language model (VLM) performs reflective evaluation, updating the current best reward design and generating a textual summary of the observed task performance (\textbf{motion-language representation}, line 5-6). incorporates both the motion description and the VLM-generated summary from the previous iteration to refine the reward generation process (\textbf{language-guided reward design}, line 7). Below we discuss the implementation details of each components for effective reward learning of musculoskeletal control.

\subsection{Musculoskeletal motion dynamics}

We evaluate each reward proposal by performing policy optimization to generate dynamical control trajectories as feedback. In \algo, we adopt \mpc~as the control policy, a model-based planner that employs a hierarchical control pipeline for musculoskeletal systems~\citep{wei2025motion}. Compared with reinforcement learning based control requiring hours to days of training, \mpc~ employs a training-free pipeline which significantly reduces the policy optimization time to minutes, allowing more reward learning iterations in \algo~(see Appendix \ref{sec:mpc2} for method details). The resulting musculoskeletal motion dynamics obtained by rolling out the optimized policy serve as the dynamical feedback for refining the reward specification.

\subsection{Motion-language representation}

We use the VLM as a semantic observer that produces interpretable, language-based evaluations rather than scalar scores. The VLM compares the rendered control dynamics $\boldsymbol{\zeta}^{(i)}$ against the dynamics generated under the current best reward definition. If the newly proposed reward yields control sequences that better align with the motion description, both the current best reward $r^*$ and the corresponding control dynamics $\boldsymbol{\zeta}^*$ are updated. The VLM also produces structured textual feedback $\mathcal{F}$ of $r^*$ which qualitatively evaluates the motion relative to $R$. This feedback serves as reflective input to the LLM, guiding subsequent iterations of reward synthesis. Through this mechanism, \algo~establishes a multimodal interface for specifying and interpreting complex motor behaviors, integrating visual and textual modalities to reason about the correspondence between natural-language motion descriptions and observed motion.

In practice, to improve robustness, motion evaluations are verified through multiple VLM assessments to reduce the impact of inconsistent feedback. When incorrect evaluations occur, resulting reward updates typically fail to improve performance and are rejected in subsequent iterations, ensuring unreliable feedback does not propagate through the refinement process.

\subsection{Language-guided reward design}

Designing executable reward functions from multimodal inputs requires mapping semantic and structural priors to physically meaningful quantities. In musculoskeletal control, this must capture nonlinear couplings between balance, posture, and coordination which are difficult to encode manually. We employ a language-guided reward synthesis process, where an LLM generates interpretable reward terms by reasoning over both linguistic and structural context:

\textbf{(1) Motion Description.} The natural-language specification $R$ defines the high-level control objective (e.g., “make the arm grasp and lift the bottle”). It serves as a semantic prior that highlights key performance factors such as grasp stability, coordination, and smoothness.

\textbf{(2) Environment.} The environment $\mathcal{E}$ specifies the state, control, and transition dynamics. Parsing $\mathcal{E}$ allows the identification of physically relevant variables (e.g., joint angles, actuator lengths, and contact forces) that can parameterize executable reward terms.

Given the contextual information and VLM-based evaluations $\mathcal{F}$, the LLM performs a local search over the current best reward function $r^*$ to synthesize new reward terms. Each term encodes a biomechanical sub-objective such as orientation tracking, smoothness, or joint stability. At each iteration, candidate rewards are evaluated through control rollouts and accepted only if they improve performance. The inherent non-determinism of LLM and VLM outputs encourages exploration across iterations, preventing the search from getting stuck in suboptimal solutions even when proposals are rejected. This process continues until an effective reward function is identified, yielding an interpretable objective suitable for policy optimization and control-based dynamical feedback.

Compared with traditional language-based approaches that rely solely on LLMs, \algo~incorporates dynamical control feedback, enabling the VLM to reflect on kinematic and postural precision, an essential capability for achieving stable musculoskeletal control.

\begin{figure}[H]
    \centering
    \includegraphics[width=\textwidth]{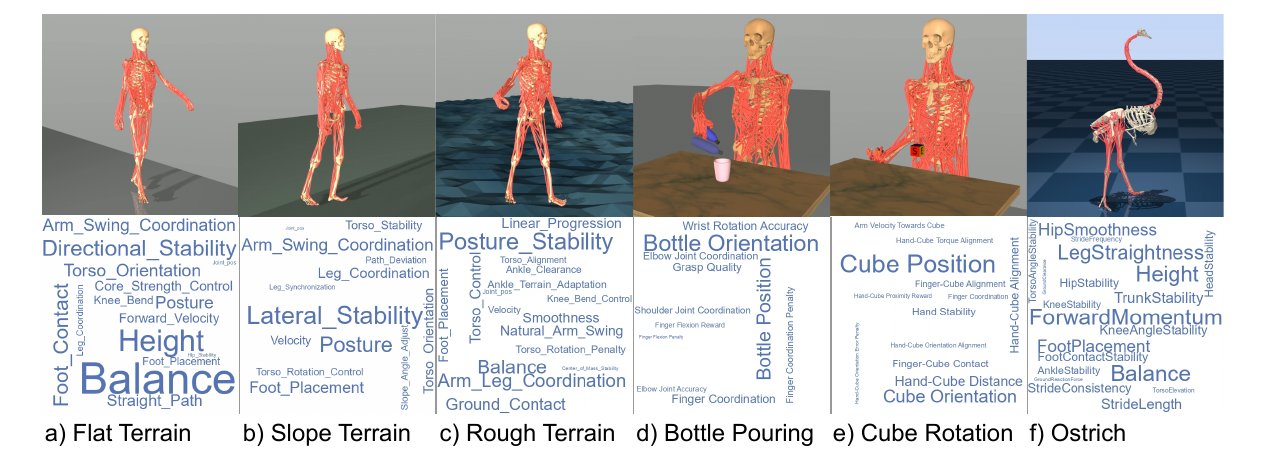}
    \vspace{-20pt}
    \caption{Overview of the six evaluated tasks. The top row illustrates the environment setup for each task, and the bottom row visualizes the relative weighting of learned reward terms.}
    \label{fig:environments}
    \vspace{-10pt}
\end{figure}

\section{Experiments}

We evaluate \algo\ across a diverse set of musculoskeletal systems and tasks, assessing its capacity to explore the space of possible reward functions, solve novel tasks, and integrate different forms of human input. Unless otherwise noted, all VLM and LLM-based reward and feedback algorithms are built on the Gemini~\citep{team2023gemini} and Qwen models~\citep{yang2024qwen2}, specifically \texttt{gemini-2.0-flash} and \texttt{Qwen2.5-Coder-32B-Instruct} models. 

\subsection{Environments}
As shown in Figure~\ref{fig:environments}, our experimental setup comprises a diverse set of musculoskeletal systems and tasks in the MuJoCo simulator~\citep{todorov2012mujoco}, capturing a broad spectrum of control challenges. The suite includes locomotion environments spanning flat, rough, and sloped terrain, testing stability and adaptability under varying conditions. Within the flat-terrain setting, we further consider directional turning and an injured-body condition where selected leg muscle groups are weakened, enabling evaluation of gait robustness and compensatory strategies. In addition to locomotion, the setup includes manipulation tasks such as bottle pouring and cube manipulation that emphasize coordination and precision, as well as a non-human locomotion task based on an ostrich muscle model to assess generalization beyond human morphology.

\subsection{Experimental Results}

\noindent \textbf{Comparison with state-of-the-art LLM/VLM based methods.}
We evaluate our method against three baselines: human-engineered reward functions \citep{wei2025motion}, Eureka~\citep{ma2023eureka}, and HARMON~\citep{jiang2024harmon} on both locomotion and manipulation tasks. Our implementation details of method and baselines are elaborated in Appendix \ref{sec:baselines_imp}. Evaluation focuses on final performance after convergence, measured by average walking distance over 10 seconds for locomotion and by object position and orientation errors for manipulation. 

\begin{figure}[thp]
    \centering

    \begin{minipage}{0.48\textwidth}
        \centering
        \includegraphics[width=\linewidth]{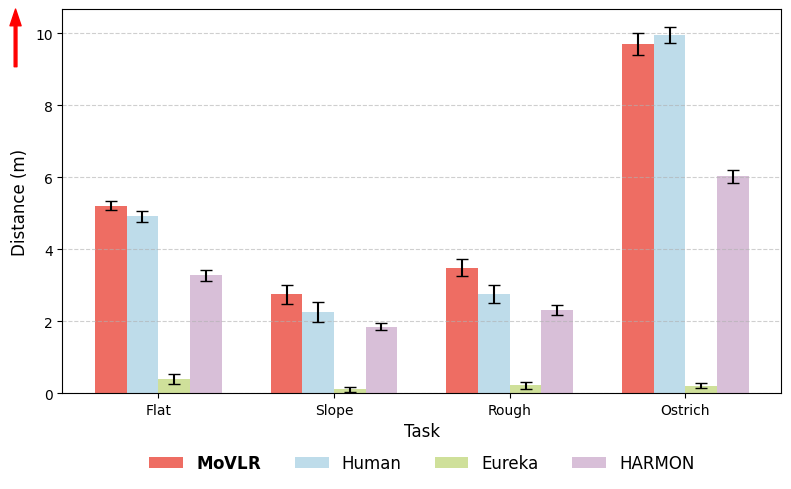}
        \caption*{(a) Locomotion Tasks}
    \end{minipage}
    \hfill
    \begin{minipage}{0.48\textwidth}
        \centering
        \includegraphics[width=\linewidth]{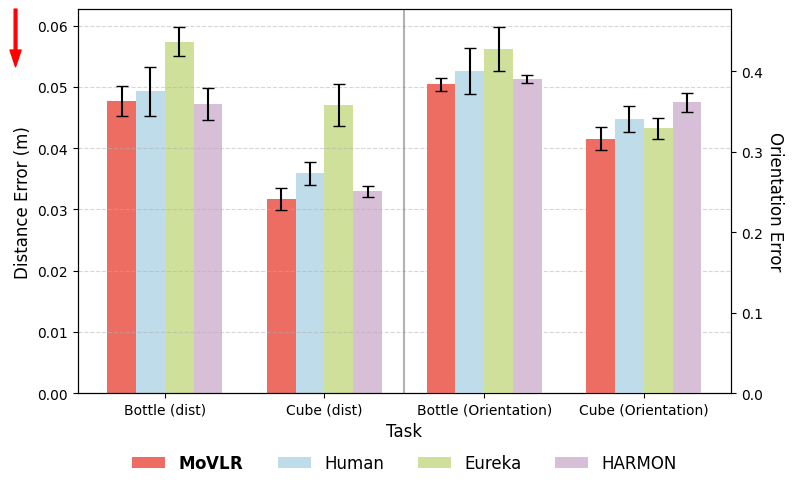}
        \caption*{(b) Manipulation Tasks}
    \end{minipage}

    \caption{Performance comparison of \algo\ against baselines across (a) locomotion and (b) manipulation tasks. Locomotion is measured by total distance walked in 10s (higher is better), while manipulation is evaluated by object distance and orientation errors (lower is better).}
    \label{fig:performance}
\end{figure}

As shown in Figure~\ref{fig:performance} (a), across all locomotion environments, \algo\ consistently achieves higher task performance, yielding the longest walking distances on flat, sloped, and rough terrains. The gains are most pronounced in challenging settings, where terrain irregularities demand adaptive stability and coordinated motion. Compared to HARMON, which relies on visual alignment over discrete frame sequences, \algo\ integrates feedback more tightly with control dynamics by using full trajectory rollouts to inform structured reward updates, enabling the capture of temporally extended dependencies such as gait rhythm, balance recovery, and joint coordination. Although \algo\ performs slightly below the human baseline, it generalizes effectively to the ostrich environment, maintaining strong performance despite significant morphological differences. These results indicate that multimodal reward refinement produces robust and transferable control objectives across biomechanical structures.

As shown in Figure~\ref{fig:performance} (b), \algo\ achieves the lowest average position and orientation errors compared to all baselines in the manipulation tasks, indicating more precise and stable object interactions. The improvements are consistent across both bottle-pouring and cube-rotation movements, suggesting that multimodal feedback enhances the alignment between high-level motion intent and low-level control behavior.

\begin{figure}[htp]
    \centering
    \includegraphics[width=\linewidth]{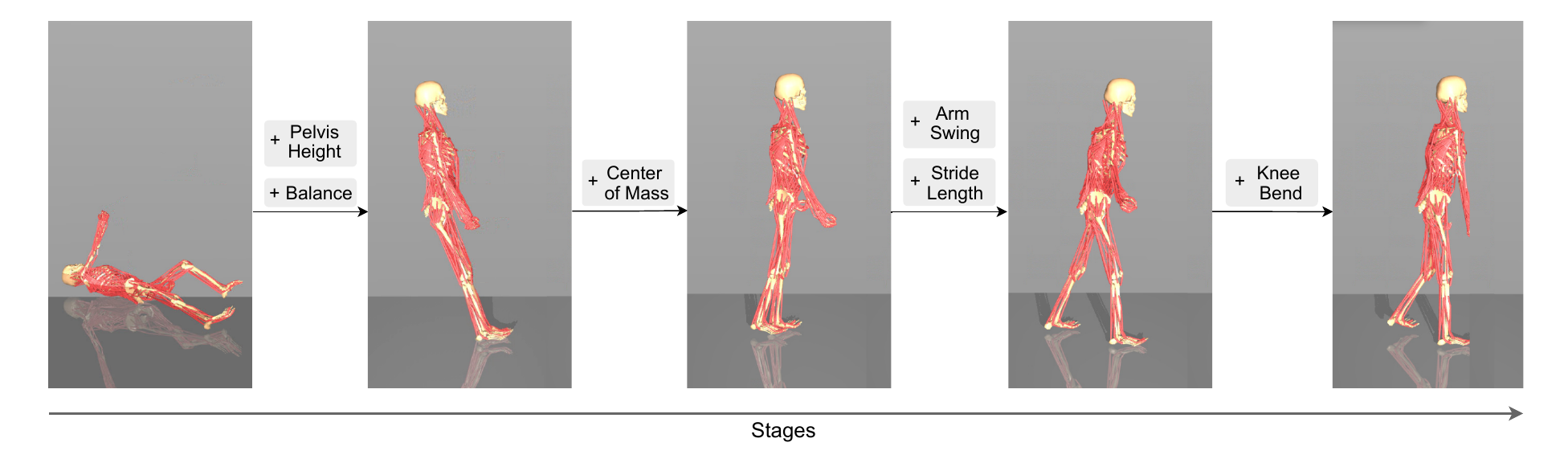}
    \vspace{-10pt}
    \caption{Progressive improvement of the musculoskeletal model’s gait across training stages based on movement video.}
    \label{fig:videofeedback}
    
\end{figure}

\noindent \textbf{Evolution of weighted reward terms across refinement stages.}
The progression of residual reward weights across refinement stages reveals how the feedback-driven process reorganizes the internal optimization landscape toward biomechanically consistent behavior. Visual inspection of the heatmaps shows clear temporal structure in how specific reward terms are emphasized, attenuated, or replaced as refinement proceeds. Rather than uniform or random variation, the weights evolve in a task-specific and interpretable manner that reflects the gradual integration of control priorities derived from feedback. This progression is visually illustrated in Figure~\ref{fig:videofeedback}, showing the musculoskeletal agent’s transition from instability to coordinated walking as successive stages refine control priorities such as balance, posture, and stride formation.

\begin{figure}[thp]
    \centering
    \begin{minipage}{0.48\textwidth}
        \centering
        \includegraphics[width=\linewidth]{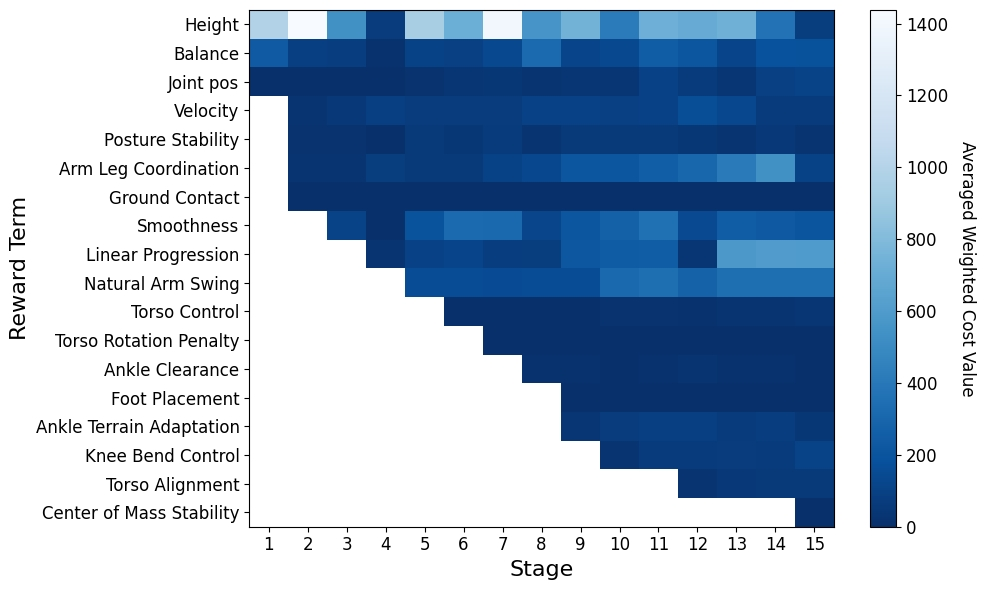}
        \caption*{(a) Rough Terrain}
    \end{minipage}
    \hfill
    \begin{minipage}{0.48\textwidth}
        \centering
        \includegraphics[width=\linewidth]{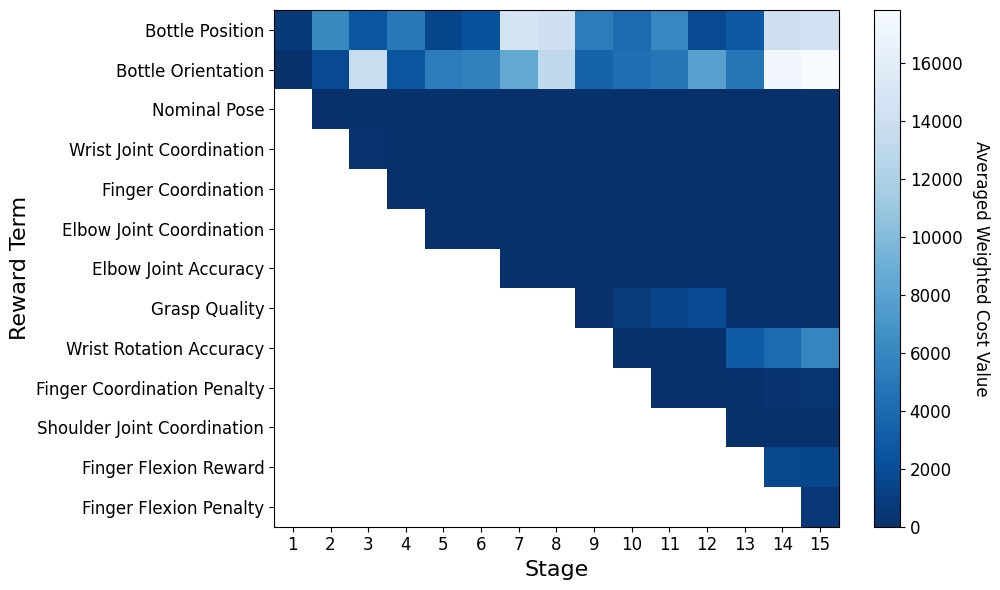}
        \caption*{(b) Bottle Pouring}
    \end{minipage}
    \caption{Weighted reward terms per stage for (a) locomotion task, (b) manipulation task}
    \label{fig:progress}
\end{figure}

In Figure~\ref{fig:progress} (a), we demonstrate the learning evolution of \algo~in locomotion task over rough terrain. We observe early refinement stages concentrate weight on coarse global stability terms such as \textit{height}, \textit{velocity}, and \textit{balance}. These initial weightings dominate the first few iterations, suggesting that the system prioritizes feasibility and upright posture before attempting finer coordination. As refinement progresses, the influence of these global terms decreases steadily, while localized biomechanical descriptors, such as \textit{foot placement}, \textit{hip alignment}, and \textit{knee control}, become more prominent. This redistribution indicates a shift from whole-body stabilization to detailed gait regulation. By later stages, the weight profiles become highly structured, with consistent activation around terms governing \textit{step symmetry}, \textit{torso orientation}, and \textit{ankle stability}, suggesting convergence toward coordinated, rhythmic locomotion.

In the bottle pouring task shown in Figure~\ref{fig:progress} (b), a similar hierarchical refinement pattern is observed. Initial stages emphasize gross spatial alignment through terms such as \textit{bottle position} and \textit{bottle orientation}, enabling task feasibility. With continued refinement, weights shift toward fine motor control components, including \textit{grasp quality, elbow joint accuracy}, and \textit{finger coordination}. The redistribution of weights toward distal control terms indicates that the framework captures the need for precise joint coordination in achieving smooth and stable object manipulation. 

Additional heatmaps for the remaining four tasks are included in appendix~\ref{sec:additional}, illustrating consistent refinement dynamics across both locomotion and manipulation domains.

\textit{Reward Interpretability.} The learned reward terms exhibit strong semantic and structural interpretability, with many feature names corresponding to well-defined biomechanical quantities that admit explicit formulations. For example, \textit{height} is often expressed as $h = z_{\text{COM}}$, the vertical position of the center of mass. Similarly \textit{forward velocity} can be represented as $v_x = \dot{x}*{\text{COM}}$, and \textit{balance} as the deviation between the projected center of mass and the support polygon, e.g., $| \Pi{\text{support}}(x*{\text{COM}}) - x_{\text{foot}} |$, etc. More generally, reward terms can be viewed as encodings of desired biomechanical properties, where these formulations provide a direct pathway to constructing executable reward functions that can be optimized within the control framework.



\noindent \textbf{Ablation Studies.} To better understand the contribution of each component and the generality of the proposed framework, we conduct a series of ablation studies examining (1) reward generalizability across environments, model conditions and policy parameterization; (2) the use of a single unified vision–language model for feedback and code generation.

\textit{Reward generalizability.} We test transferring reward functions proposed on flat terrain to new environments without additional refinement. The transferred rewards show strong generalization across terrains and morphologies. While performance drops moderately on rough (2.41 m vs. 2.76 m) and sloped (1.99 m vs. 2.25 m) terrains, agents remain stable and capable of sustained locomotion. In the injured-body setting, transfer performance slightly improves (5.12 m vs. 4.8 m), indicating robustness to actuator failure. The method also enables a left-turn behavior previously infeasible with hand-engineered rewards, showing that the learned reward structure extends beyond the original environment configuration. 

We further test reward functions learned by \algo\ in a reinforcement learning setting using DynSyn~\citep{he2024dynsyn} for the bottle-pouring task. The transferred rewards enable successful task completion without further tuning, producing stable pouring trajectories, demonstrating that \algo-designed rewards capture generalizable structure transferable across control algorithms.

\textit{VLM-only reward learning.} The framework’s design separates the vision–language feedback and the language-based code generation components. To assess whether this modularity is necessary, we implement a unified configuration in which a single multimodal model performs both feedback interpretation and code synthesis. The unified variant shows substantially degraded performance, often producing invalid or incomplete reward code and failing to improve control behavior over iterations. These observations suggest that current vision–language models do not yet possess the compositional or programmatic reasoning required to perform both tasks simultaneously, highlighting the importance of maintaining distinct feedback and synthesis stages.

\section{Conclusion and Discussion}
In this work, we introduce \algo~, an automatic workflow that leverages vision-language models to bridge explicit language descriptions with implicit motor control required for high-dimensional musculoskeletal systems. By integrating multimodal feedback into the learning loop, \algo\ finds biomechanically grounded reward functions that are iteratively refined to guide the musculoskeletal agent toward stable, natural motion. Through this method, we demonstrate that VLMs can successfully translate high-level motion descriptions into detailed control objectives, improving musculoskeletal performance across diverse environments and tasks.

Experimental results across locomotion and manipulation tasks show that \algo\ consistently outperforms both human-designed and language-based baselines. The iterative refinement of rewards mirrors the hierarchical structure of human motor learning, progressing from coarse stability constraints to fine-grained joint coordination.

Beyond performance gains, \algo\ highlights a fundamental connection between \textit{explicit language intent} and \textit{implicit reward emergence}. Acting as a perceptual bridge, the VLM grounds linguistic goals in physical dynamics, producing interpretable and dynamically consistent reward representations. This work offers a scalable and principled path toward biologically plausible, interpretable, and generalizable control for complex behaviors and morphologies.

\acks{This work is supported by STI 2030-Major Projects 2022ZD0209400 and NSFC 62461160313. Correspondence to: Yanan Sui (ysui@tsinghua.edu.cn).}

\bibliography{references}

\newpage
\appendix
\section{}

\subsection{Model Predictive Control with Morphology-Aware Proportional Control}\label{sec:mpc2}
Model Predictive Control with Morphology-Aware Proportional Control (\mpc)~\citep{wei2025motion} is a hierarchical control scheme for high-dimensional musculoskeletal systems. 
Let $z \in \mathbb{R}^{d_z}$ denote the major joint coordinates defining the system posture ($d_z \ll d_u$, where $d_u$ is the actuator dimension). 
The high-level planner solves
\begin{equation}
z^* = \arg\min_{z} \sum_{h=0}^{H-1} C(s_{t+h}, u_{t+h}), 
\quad u_{t+h} = \pi_{\mathrm{MP}}(s_{t+h}, z),
\end{equation}
using a sampling-based MPC (e.g., MPPI) over posture space. 
Instant rollouts are introduced by sampling candidate $z$ around the current posture $M_{\mathrm{pos}}(s_t)$ for rapid recovery from disturbances.

The low-level morphology-aware proportional controller maps the target posture $z^*$ to target actuator lengths $l^*$ and computes desired actuator forces
\begin{align}
    f_m^* = \min\big(0,\, K \cdot (l^* - l)\big), \quad K = \bar{k} \sum_{i \in I_z} \left| \mathrm{col}_i(J_m) \cdot \big(z^*_i - M_{\mathrm{pos}}(s_t)_i\big) \right|,, 
\end{align}

where $K$ is the proportional gain of actuators, and $\bar{k}$ is the global scalar parameter. 
Actuator commands $u^*$ follow from first-order actuator dynamics. This decomposition reduces the optimization dimension from $H \cdot d_u$ to $d_z$, enabling zero-shot control across morphologies without training. 

\subsection{Baseline Methods}\label{sec:baselines_imp}

\textbf{Eureka}~\citep{ma2023eureka} We adapt the Eureka framework, which uses large language models to synthesize reward functions from textual motion descriptions. For fair comparison, we implement Eureka using the same closed-loop setting as our method, but without the vision-language feedback: the language model receives textual summaries of agent rollouts rather than video-based feedback. The number of optimization rounds, samples per round, and other training parameters are matched to our method to ensure a controlled comparison.

\noindent\textbf{HARMON}~\citep{jiang2024harmon} We adapt the HARMON framework, which combines large language model reasoning with visual motion priors to generate whole-body humanoid motions. For fair comparison, we employ HARMON in our musculoskeletal control setting by using the same closed loop setting as our method, but replacing the video feedback with image feedback: the VLM receives 4 evenly spaced frames extracted from the rendered video rather than the full video. The number of optimization rounds, samples per round, and other training parameters are matched to our method to ensure a controlled comparison.

\noindent\textbf{Human} We use hand-crafted reward functions provided with the musculoskeletal tasks as a baseline. These rewards are designed by domain experts and encode task objectives through manually specified heuristics~\citep{wei2025motion}. 

\subsection{Additional discussion on Runtime, Inference Cost, and Scalability}

We provide a discussion on the computational characteristics of our method in terms of runtime, inference cost, and scalability. A typical run of the framework involves multiple iterative stages of policy optimization, video rendering, and VLM/LLM inference. In our experiments, we use up to 15 stages, although strong performance can often be achieved within 5-10 stages. Per stage, policy optimization requires approximately 5–6 minutes, video rendering ~1 minute, and each VLM/LLM inference ~1 minute, yielding a total of ~10 minutes per stage. This corresponds to ~35–40 minutes for a 5-stage run and ~105–120 minutes for a full 15-stage run. In terms of inference cost, each run consumes approximately 30,000 tokens (combined video and language input), with the exact monetary cost depending on the specific VLM/LLM used. Overall, the framework scales linearly with the number of stages and can be adjusted to trade off runtime and performance.

\subsection{Prompts and Examples}

\begin{figure}[htp]
    \centering
    \includegraphics[width=\textwidth]{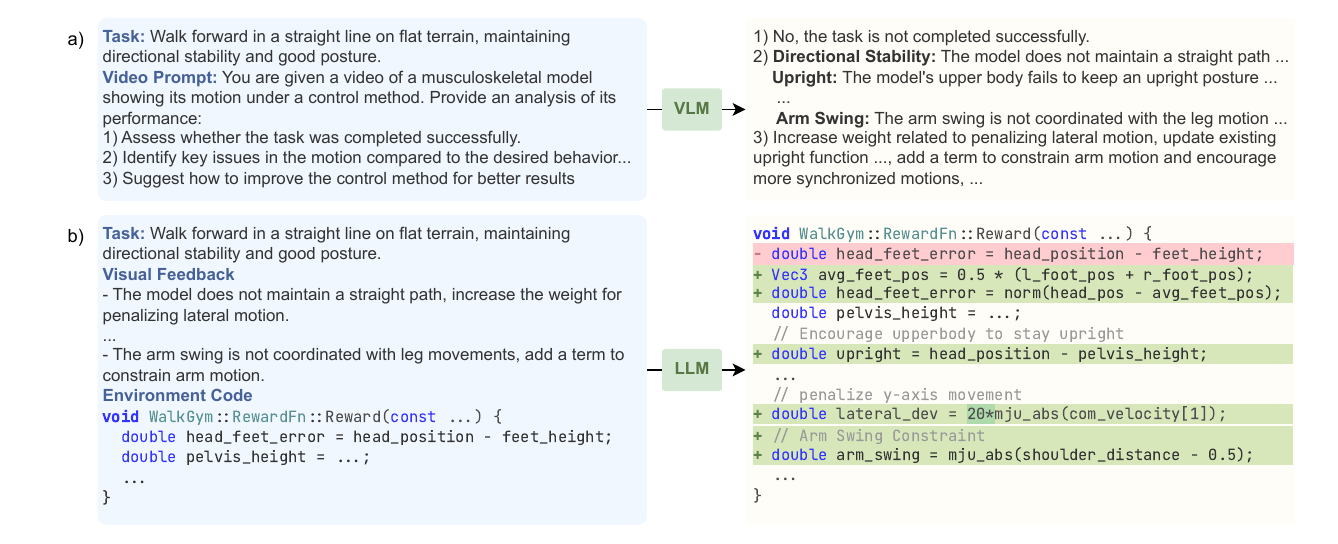}
    \caption{Example inputs and outputs of the (a) VLM, and (b) LLM. The VLM analyzes a video motion sequence based on the given motion description and provides diagnostic feedback. The LLM uses this feedback to explore corresponding code modifications to the reward function.}
    \label{fig:promptexamples}
\end{figure}

Throughout our pipeline, we use multiple prompts to guide both perception and reward synthesis. The primary prompts correspond to (i) motion–language representation, where the VLM is prompted to evaluate rendered motion trajectories against a task description and produce structured feedback, and (ii) language-guided reward design, where the LLM is prompted to refine reward functions based on the task specification, environment context, and VLM feedback. Brief examples of these prompts are illustrated in Figure~\ref{fig:promptexamples}, while detailed, concrete prompt instances used in our experiments are provided below.

\textbf{Visual Feedback Prompt}

\begin{lstlisting}
You are given a video of a {body} muscle skeleton model whose task is to {task}. The video shows the model's actions after training a reinforcement learning model. The goal is to perform the task well and correctly. Provide a detailed critical analysis of the model's performance. You should be critial and point out specifically areas that need to be improved. Provide your analysis in a clear and concise manner, using appropriate technical language and terminology where necessary.

The reward terms used to train the reinforcement learning model, including their weights, are listed below. 

{reward_terms}

Please perform the following analysis:

a) First determine whether the task {task} was completed successfully, answer YES or NO.
b) Identify the main issues with the motion produced compared to the desired motion from the task description. First focus on successfully completing the general task, then fine-tuning details. If the task is not successfully completed yet do not worry about fine-tuning details. Be detailed with descriptions. Also analyze what specific motions in the video could cause the issues or failures. Focus mainly on {focus} motion/issues. Describe directions from the point of the view of the muscle skeleton rather than a third person view.
c) Someone is trying to run a control method to perform better than what was shown in the video, and needs some suggestions about some reward terms that could be used, added, or given greater/less weight. For new terms, assign a reasonable weight value between 0 and the maximum weight, and increase/decrease gradually if/when necessary. Do not suggest too many or redundant terms. If suggesting a new term, also suggest how the function should be defined (using words is enough, don't need to use specific functions/coding names). Given the video, issues, and existing reward terms listed above, provide some suggestions.

Be specific in your observations and suggestions. Your goal is to help improve both the correctness and the naturalness of the {body}'s motion
\end{lstlisting}

\noindent \textbf{Coding Language Model Prompt}

\begin{lstlisting}
You are updating the residual function of a MuJoCo muscle-skeleton environment using a **conservative, feedback-driven edit policy** to improve the performance of the task.

**Inputs**
- Goal/task: {task}
- Environment code (contains residual function):
$[$env_code$]$
{env_code}
$[$env_code$]$
- Task file (canonical list of valid sensors):
$[$task_code$]$
{task_code}
$[$task_code$]$
- The weights for each residual term during previous stages are provided below.
{residual_terms}
- Video feedback after analyzing a single round of running the muscle skeleton performing the task:
{feedback_string}

**Editing guidelines**
- Make a **small number of localized changes** that directly address issues observed in the feedback.
- When possible, prefer adjusting existing residual terms (e.g., scaling, weighting, or tuning) before introducing new ones.
- New residual terms may be added if they clearly align with the feedback and are supported by the task file.
- Residual functions should be defined carefully and with enough detail
- Keep edits focused on the relevant regions; avoid broad or unrelated modifications.
- Do not include weight term implementations in the environment code, all terms should be multiplied by 1.
- Pay attention to comments in the code if they exist in the code
- Ensure that the residual function remains stable and interpretable across training stages.
- When editing the residual function, the following vector/quaternion operations can be used

{operations}

{code_tips}

**Output format (strict)**
- Output the **entire, updated environment code** in a single ```cpp``` block.
- No explanations, no diffs, no comments, only the final code.
\end{lstlisting}

\noindent \textbf{Selection}

\begin{lstlisting}
You are an expert biomechanical analyst. You will be shown two videos, each depicting a muscle-skeleton model performing the task {task}. Carefully observe both performances and compare how accurately, smoothly, and efficiently the models complete the task.

Evaluate each video based on key biomechanical factors: task success, balance and stability, posture and alignment, joint coordination, and overall movement naturalness. Consider whether the motion looks physically plausible and efficient, without unnecessary or unstable compensations. Pay attention to gait or limb symmetry, center-of-mass control, and the sequencing of major joints.

After analyzing both videos, choose which one demonstrates better completion of the task, that is, which looks more correct, natural, stable, and biomechanically efficient.
Respond with only one of the following words: "first" or "second", followed by a brief explanation justifying your choice.
\end{lstlisting}

\noindent \textbf{Task Descriptions}

\noindent Flat Terrain
\begin{lstlisting}
Walk forward in a straight line on flat terrain at a velocity of about 1 m/s, maintaining directional stability and good posture
\end{lstlisting}

\noindent Slope Terrain
\begin{lstlisting}
Walk forward in a straight line on sloped terrain at a velocity of about 1 m/s, maintaining directional stability and good posture
\end{lstlisting}

\noindent Rough Terrain
\begin{lstlisting}
Walk forward in a straight line on rough terrain at a velocity of about 1 m/s, maintaining directional stability and good posture
\end{lstlisting}

\noindent Bottle Pouring
\begin{lstlisting}
Grasp and reorient the darker-shaded bottle to match the target orientation and position, indicated by a lighter shade bottle
\end{lstlisting}

\noindent Cube Rotation
\begin{lstlisting}
Grasp and reorient the cube to match the target orientation, keeping the cube held approximately in front of the musculoskeleton's chest
\end{lstlisting}

\noindent Ostrich
\begin{lstlisting}
Make an ostrich walk forward in a straight line on a flat terrain with velocity approximately 1 m/s and proper gait and posture (flat body, relatively straight legs, stable head)
\end{lstlisting}

\noindent Injured Body
\begin{lstlisting}
Make a human muscle skeleton model with right-side injuries to the biceps, gastrocnemius, semimembranosus, and semitendinosus muscles walk forward in a straight line on a flat terrain
\end{lstlisting}

\noindent Left-Turn
\begin{lstlisting}
Walk forward with good posture, then make a left turn and walk towards the new facing direction after making the turn
\end{lstlisting}

\subsection{Additional Experimental Results}
\label{sec:additional}

\textbf{Evolution of weighted reward terms for remaining tasks}

We provide supplementary heatmaps visualizing the evolution of reward term weights across refinement stages for the remaining four tasks: flat terrain, slope terrain, ostrich locomotion, and cube rotation. 

\begin{figure}[H]
    \centering
    \begin{minipage}{0.48\textwidth}
        \centering
        \includegraphics[width=\linewidth]{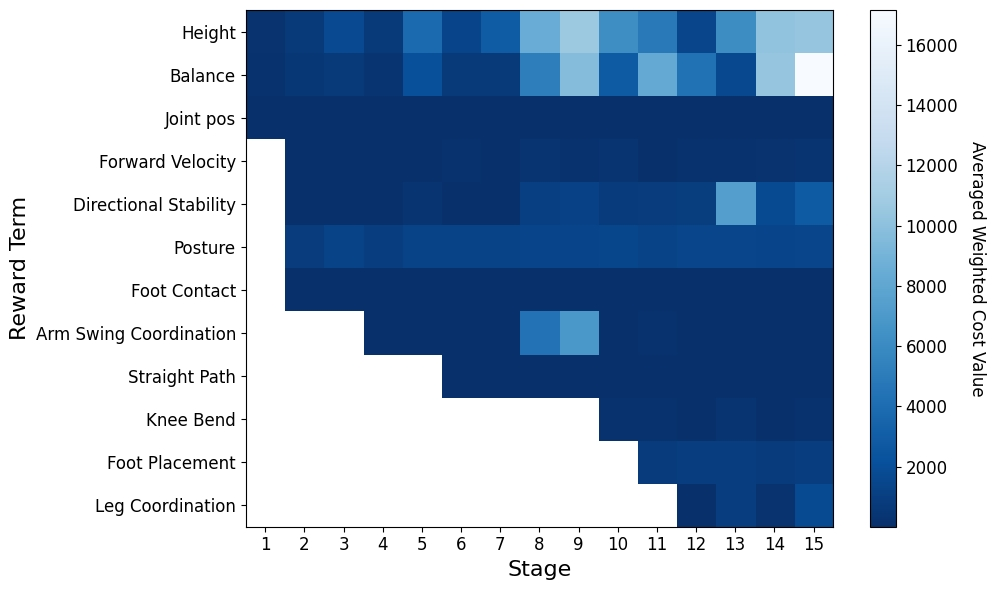}
        \caption*{(a) Flat Terrain}
    \end{minipage}
    \hfill
    \begin{minipage}{0.48\textwidth}
        \centering
        \includegraphics[width=\linewidth]{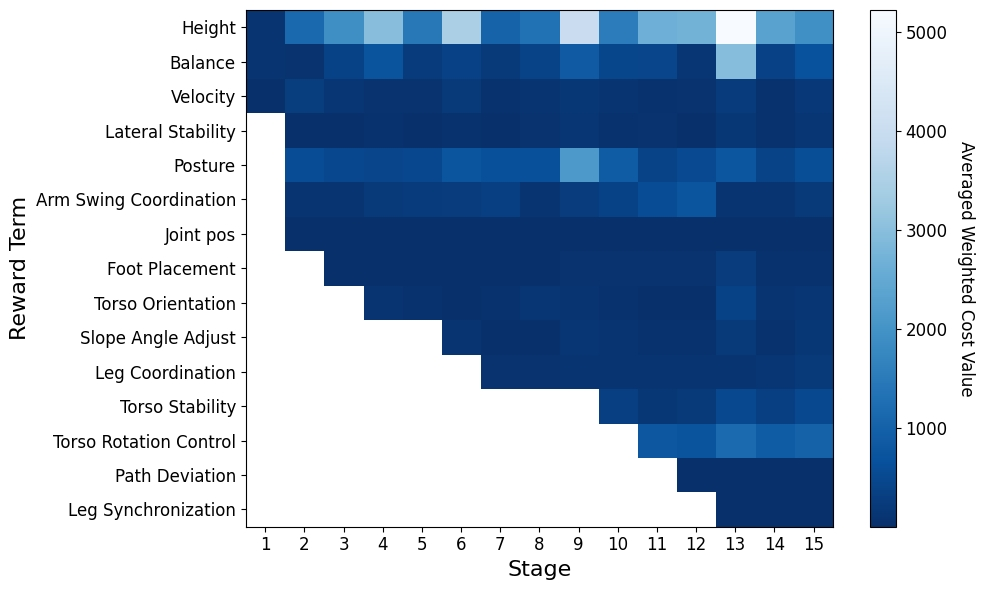}
        \caption*{(b) Slope Terrain}
    \end{minipage}
\end{figure}
\begin{figure}[H]
    \centering
    \begin{minipage}{0.48\textwidth}
        \centering
        \includegraphics[width=\linewidth]{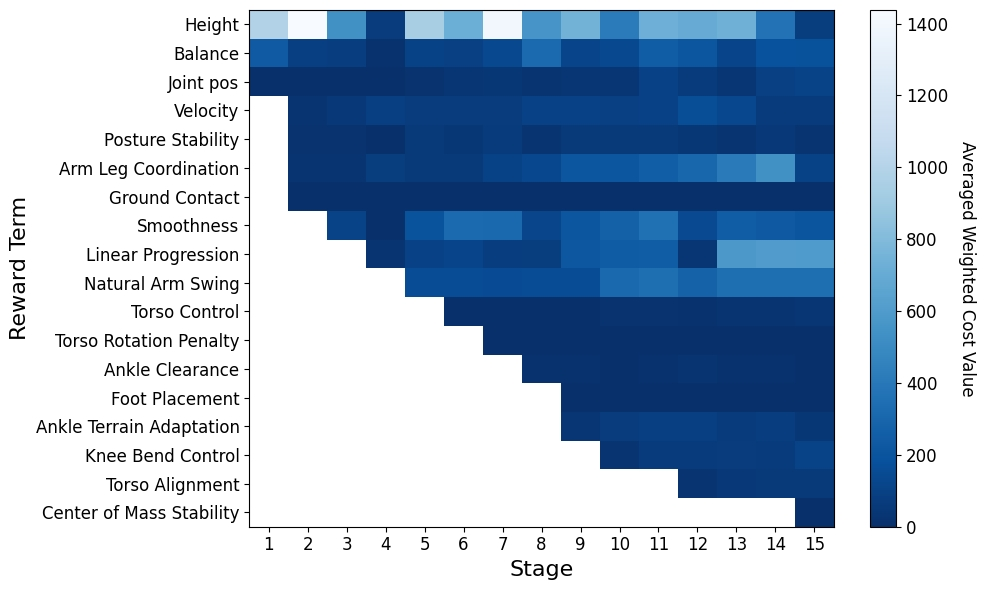}
        \caption*{(c) Ostrich}
    \end{minipage}
    \hfill
    \begin{minipage}{0.48\textwidth}
        \centering
        \includegraphics[width=\linewidth]{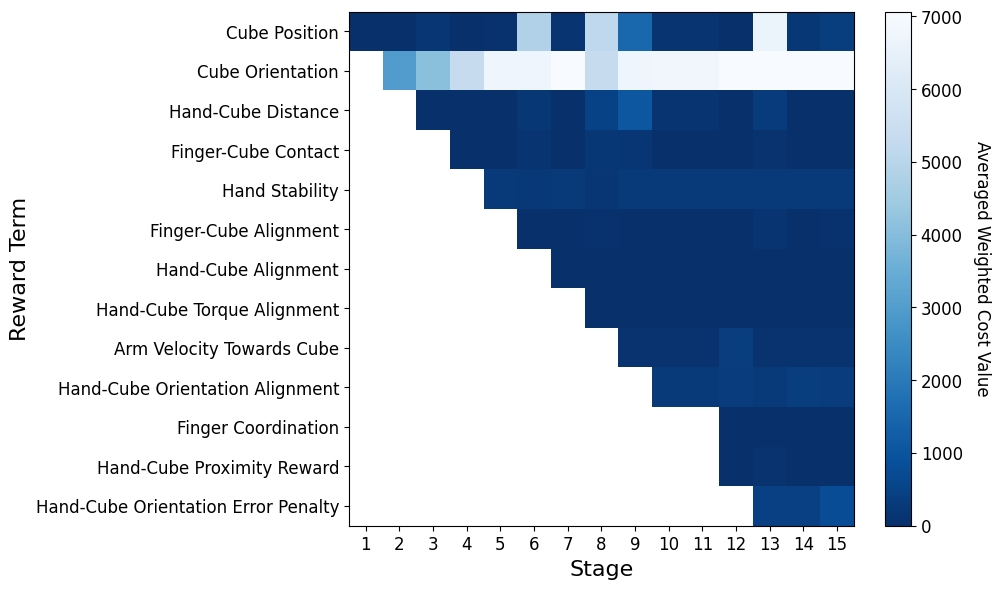}
        \caption*{(d) Cube Rotation}
    \end{minipage}
    \caption{Weighted reward terms for (a) flat terrain, (b) slope terrain, (c) ostrich, (d) cube rotation}
\end{figure}

\noindent \textbf{Comparison of Language-Designed and Human-Defined Reward Terms}

\begin{figure}[!htbp]
    \centering
    \includegraphics[width=\linewidth]{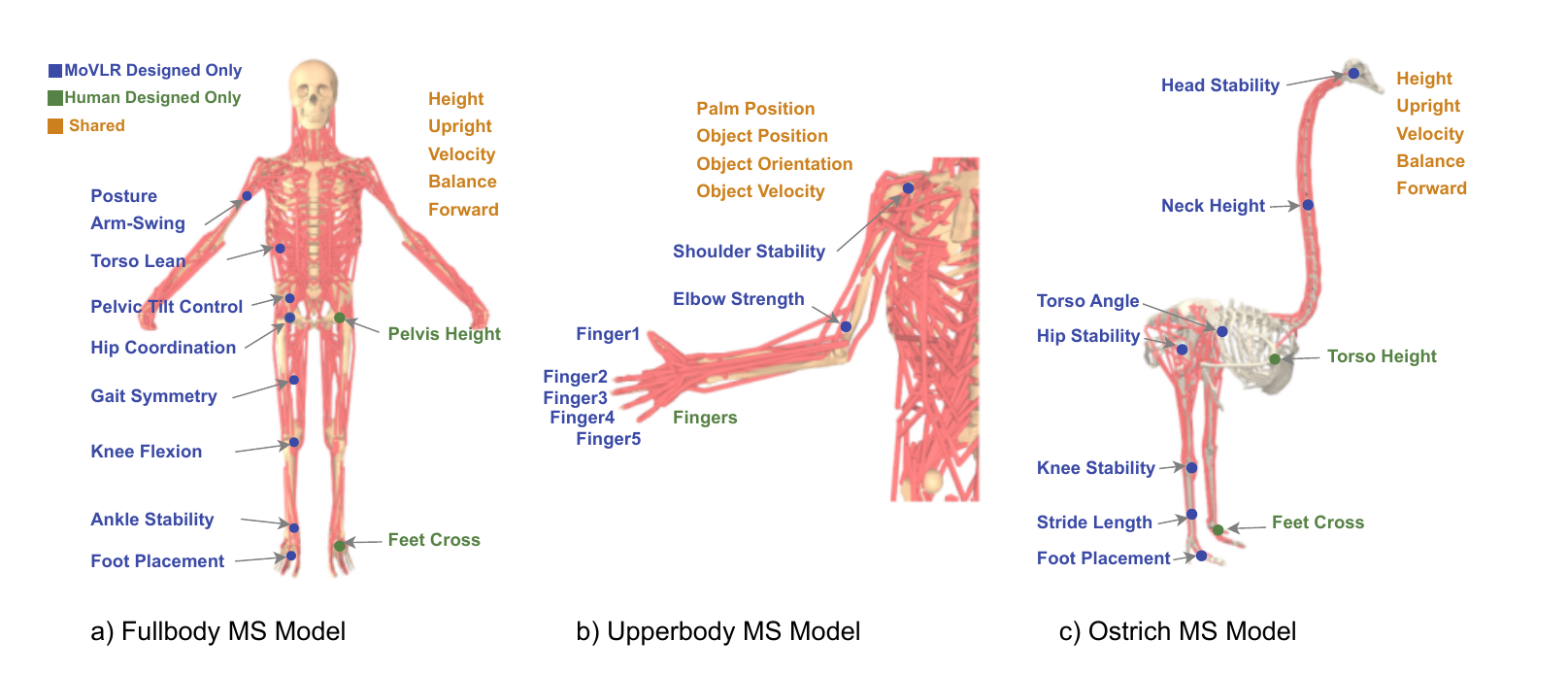}
    \caption{Comparison of reward terms designed by LLM only (blue) and by human experts (green), with shared terms shown in orange, across three musculoskeletal systems.}
    \label{fig:term_comparison}
\end{figure}

Figure~\ref{fig:term_comparison} compares the residual reward terms designed by the language model with those manually specified by human experts across three musculoskeletal systems. The comparison highlights the model's capacity to infer a more comprehensive and morphology-aware set of control objectives. 

In the fullbody model, the language model introduces a broader range of biomechanically grounded terms -- such as \textit{pelvis tilt control, hip coordination}, and \textit{gait symmetry} -- which extend beyond the coarse global stability terms (\textit{height, velocity, balance}) typically defined by human experts. For the upperbody model, the model captures fine-grained kinematic relations including \textit{elbow strength, wrist rotation}, and per-finger coordination, reflecting an understanding of localized control relevant to manipulation tasks. Finally in the ostrich model, the model adapts to non-human morphology with terms such as \textit{neck height, torso angle}, and \textit{head stability}, indicating a morphological generalization beyond human-centered priors.

\end{document}